\documentclass[10pt,journal]{IEEEtran}

\usepackage{comment}

\ifCLASSOPTIONcompsoc
  \usepackage[nocompress]{cite}
\else
  \usepackage{cite}
\fi

\ifCLASSINFOpdf
  \usepackage[pdftex]{graphicx}
  \graphicspath{{../images/}}
  \DeclareGraphicsExtensions{.pdf,.jpeg,.png,.eps}
  \usepackage[outdir=../images/converted/]{epstopdf}
\else
  \usepackage[dvips]{graphicx}
  \graphicspath{{../images/}}
  \DeclareGraphicsExtensions{.eps.,jpeg,.png}
\fi
\usepackage[cmex10]{amsmath}
\usepackage{amssymb}
\usepackage{bm}
\usepackage{mathtools}
\usepackage{nicefrac}
\usepackage{amsmath}
\usepackage{microtype}

\usepackage{acronym}

\ifCLASSOPTIONcompsoc
  \usepackage[caption=false,font=normalsize,labelfont=sf,textfont=sf]{subfig}
\else
  \usepackage[caption=false,font=footnotesize]{subfig}
\fi

\usepackage{url}
\usepackage{color}

\usepackage{tabularx}
\usepackage{multirow}
\usepackage{booktabs}
\usepackage{siunitx}

\newcommand{\vect}[1]{\bm{#1}}

\makeatletter
\newcommand\footnoteref[1]{\protected@xdef\@thefnmark{\ref{#1}}\@footnotemark}
\makeatother

\begin{document}

\title{Recurrent Encoder-Decoder Networks for Vessel Trajectory Prediction with Uncertainty Estimation}

\author{Samuele~Capobianco,
        Nicola~Forti,
        Leonardo~M.~Millefiori,~\IEEEmembership{Member,~IEEE,}
        Paolo~Braca,~\IEEEmembership{Senior~Member,~IEEE}
        and~Peter Willett,~\IEEEmembership{Fellow,~IEEE}%
\thanks{S. Capobianco was with the NATO Science and Technology Organization (STO) Centre for Maritime Research and Experimentation (CMRE), 19126 La Spezia, Italy. E-mail: research@samuelecapobianco.com.}%
\thanks{N. Forti, L. M. Millefiori, and P. Braca are with the NATO STO CMRE, 19126 La Spezia, Italy. E-mail: nicola.forti@cmre.nato.int, leonardo.millefiori@cmre.nato.int,
paolo.braca@cmre.nato.int. }%

\thanks{P. Willett is with the Department of Electrical and Computer Engineering, University of Connecticut, Storrs, 06269, CT, USA. E-mail: peter.willett@uconn.edu.}%
\thanks{This work is supported by the NATO Allied Command Transformation (ACT) via the project ``Data Knowledge Operational Effectiveness'' (DKOE).}%
}

\maketitle

\begin{abstract}
Recent deep learning methods for vessel trajectory prediction
are able to learn complex maritime patterns from
historical 
Automatic Identification System (AIS) 
data 
and accurately predict sequences of future vessel positions with a prediction horizon of several hours.
However, in maritime surveillance applications,
reliably quantifying the prediction uncertainty can be as important as obtaining high accuracy.
This paper extends deep learning frameworks 
for trajectory prediction tasks
by exploring 
how recurrent encoder-decoder neural networks can be tasked not only to predict but also to yield a corresponding prediction uncertainty via Bayesian modeling of epistemic and aleatoric uncertainties.
We compare the prediction performance of
two  different models based on labeled or unlabeled input data
to highlight how uncertainty quantification and accuracy can be improved by 
using, if  available, additional information on the  intention of the ship (e.g., its planned destination).
\end{abstract}

\section{Introduction}
\label{sec:intro}


Trajectory prediction -- for collision avoidance, anomaly detection and risk assessment -- is a crucial functional component of intelligent maritime surveillance systems and next-generation autonomous ships. 
Maritime surveillance systems are increasingly relying on the huge amount of data 
made available by terrestrial and satellite networks of Automatic Identification System (AIS) receivers. 
The availability of maritime big data enables the automatic extraction of spatial-temporal mobility patterns that can be processed 
by modern deep learning networks 
to enhance trajectory forecasting.

Today, most commercial systems primarily rely on 
trajectory prediction methods based on the Nearly Constant Velocity (NCV) model, since this linear model is simple, fast, and of practical operability
to perform short-term predictions of straight-line trajectories~\cite{survey19}.
However, the NCV model tends to overestimate the actual uncertainty as the prediction horizon increases.
A novel linear model, based on the Ornstein-Uhlenbeck (OU) stochastic process, 
is becoming recognized as a reliable means to improve long-term predictions~\cite{Millefiori2016}, with special focus on uncertainty reduction via estimation of current navigation settings.

\begin{figure}[t!]
	\centering%
	\includegraphics[trim=80 10 80 10,clip,width=\columnwidth]{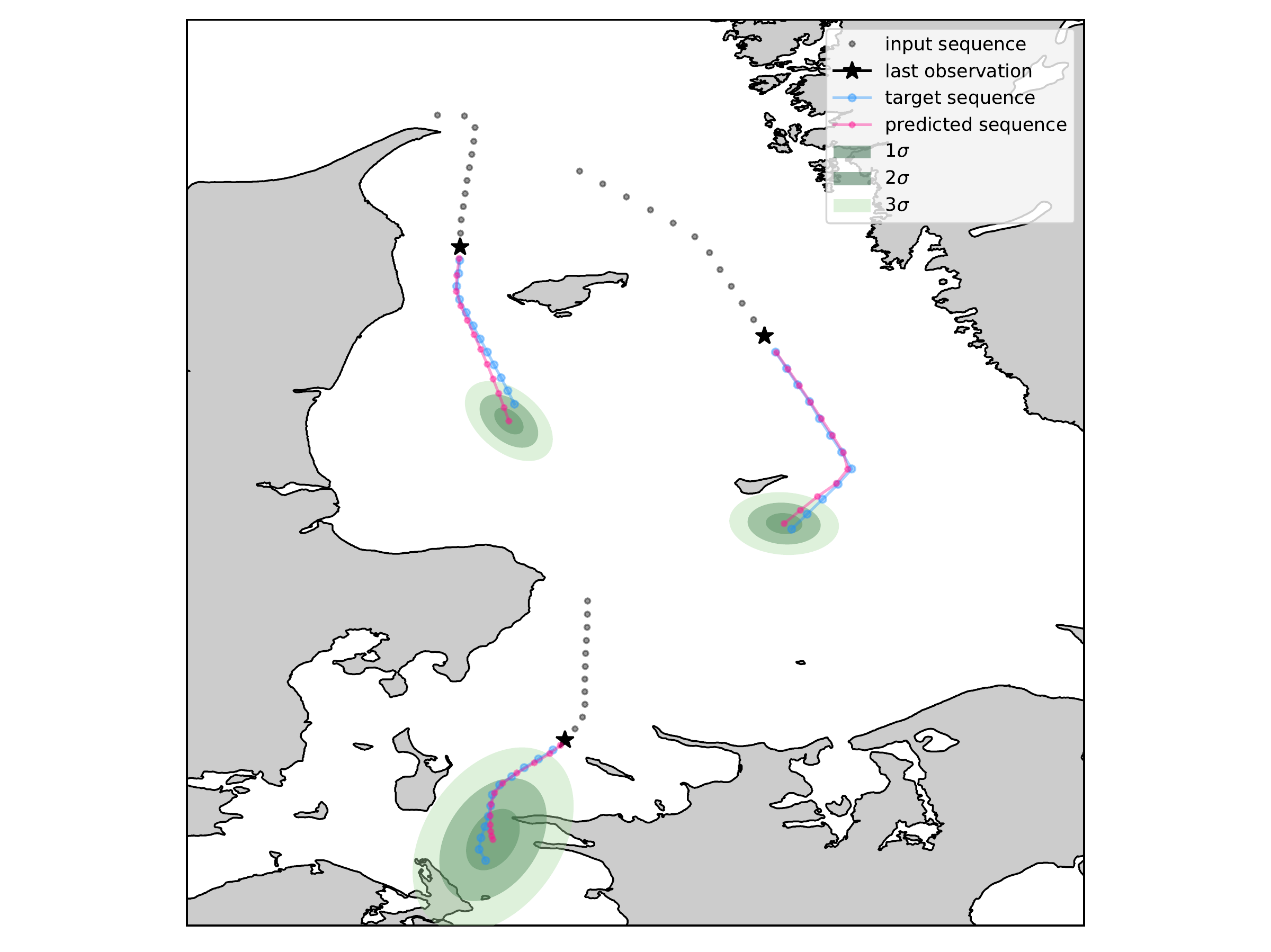}%
	\caption{Vessel trajectory prediction with representation of the prediction uncertainty depicted as confidence ellipses (at different confidence levels). Contains data from the Danish Maritime Authority that is used in accordance with the conditions for the use of Danish public data~\cite{DMA}.}
	\label{fig:pred_unc}
\end{figure}

Although most maritime traffic is very regular, 
and thus model-based methods can be easily applied, 
in the presence of 
maneuvering behaviors of the ship such models will tend to 
lack the desired prediction accuracy. 
In such cases, nonlinear and data-driven methods including
adaptive kernel density estimation~\cite{Ristic08},
nonlinear filtering~\cite{Perera2012,Mazzarella15},
nearest-neighbor search methods~\cite{Hexeberg17,Dalsnes2018},
and machine learning techniques~\cite{Zissis2017},
may provide 
more suitable solutions.
Furthermore, the latest advances in deep learning-based predictive models and the combined availability of large volumes of AIS data 
allow for enhanced vessel trajectory prediction and maritime surveillance.   
Recent approaches based on deep learning 
are documented in~\cite{Nguyen2018,Nguyen2018b,Yu2020,Zhou2020,Murray2020,Murray2021,Forti2020,Capobianco2021}.
However, standard deep learning models 
cannot provide 
predictive uncertainty,
hence no quantification of the confidence with which the prediction outputs can be trusted is available.

To accompany results from deep learning by their associated confidence levels has recently arisen to be of heightened importance, and naturally there has been significant research attention toward that goal. Data-driven methods for uncertainty quantification have been recently proposed to estimate model and data uncertainty based on ensemble~\cite{Laks2017}
or Bayesian~\cite{Gal2016,Kendall2017,Bhatta2018,Xiao2019,KuleshovFE18}
learning.
Bayesian deep learning methods apply Bayesian modeling and
variational inference to neural networks, leading to Bayesian neural networks (BNNs) that treat the network parameters as random variables
instead of deterministic unknowns
to represent the model uncertainty on its predictions.
BNNs can capture the uncertainty within the learning model, while maintaining the flexibility of deep neural networks, and hence they are particularly appealing
for safety-critical applications (e.g., autonomous transportation systems, robotics, medical and space systems)
where uncertainty estimates can be propagated in an end-to-end neural architecture 
to enable improved decision making.
In particular,~\cite{Gal2016} shows that \emph{dropout}, a well-known regularization technique to prevent neural network over-fitting~\cite{Srivastava14},
can be used as a variational Bayesian approximation of the predictive uncertainty 
in existing deep learning models trained with standard dropout,
by performing Monte Carlo (MC) sampling with dropout at test time,
i.e., by sampling the network with random omission of 
units representing MC samples obtained from the posterior distribution
over models.
In~\cite{Kendall2017} a Bayesian deep learning approach for the combined quantification of both data and model uncertainty, extracted from BNNs, is proposed for computer vision applications.
In addition, 
a neural network architecture based on Long Short-Term Memory (LSTM) with uncertainty modeling is proposed in~\cite{Jung2020}
to incorporate non-Markovian dynamic models in the prediction step of a standard Kalman filter for target tracking. 

In this paper, we present a vessel trajectory learning and prediction framework to generate future vessel trajectory samples
given the sequence of the latest AIS observations. 
The proposed method is built upon the LSTM encoder-decoder architecture presented in~\cite{Forti2020,Capobianco2021}, which has emerged as an effective and scalable model for sequence-to-sequence learning of maritime trajectories.
We extend~\cite{Forti2020,Capobianco2021} by providing a practical quantification of the predictive uncertainty via Bayesian learning tools based on MC dropout~\cite{Gal2016}. 
Preliminary work on trajectory prediction with uncertainty quantification applied to the maritime domain
was presented in \cite{fusion2021}. 

In this work, we extend \cite{fusion2021} by
providing a comprehensive description of the prediction uncertainty modeling, a detailed introduction of the variational LSTM model used to implement our encoder-decoder architecture, 
and by proposing an alternative decoder scheme with a novel regularization method of the information about the vessel's intention, which may be available in the maritime surveillance data.
Moreover, we present novel results on the estimated predicted variance, and 
a performance comparison demonstrating the gain in state prediction accuracy with respect to \cite{fusion2021}.
Fig.~\ref{fig:pred_unc} 
shows an example of how the proposed model is able to predict future trajectories and prediction uncertainty given input sequences from past data.

To summarize, the main contributions of this work are:
\begin{enumerate}
    \item A model for the aleatoric and epistemic uncertainty of trajectory predictions 
    provided by encoder-decoder RNNs 
    using Bayesian deep learning tools;
    \item A novel regularization method for the decoding phase to prevent complex co-adaptations between the encoded data and the high-level information about the vessel’s intention;
    \item Experimental results on real-world AIS data showing the effectiveness of the proposed encoder-decoder architecture with uncertainty modeling in learning trajectory predictions with well-quantified uncertainty estimates using labeled and unlabeled data.
\end{enumerate}

The paper is organized as follows. In Section~\ref{sec:problem}, we introduce the vessel trajectory prediction problem. 
In Section~\ref{sec:encdec}, 
we present the proposed attention-based encoder-decoder framework for trajectory prediction with uncertainty.
Experimental results on a real-world AIS dataset are presented and discussed in Section~\ref{sec:experiments}. Finally, we conclude the paper in Section~\ref{sec:discussion}.

\section{Problem formulation}
\label{sec:problem}

We formulate the vessel trajectory prediction problem 
as a supervised learning process 
by following a sequence-to-sequence deep learning approach
to directly generate 
the distribution of an
output sequence of future states 
given an input sequence of past AIS observations.
From a probabilistic perspective, the objective is to 
determine the following predictive distribution
\begin{equation}\label{eq:predictive_distr}
p(\mathbf{y}_{1:h}^* | \mathbf{x}_{1:\ell}^*, \mathcal{D} ),
\end{equation} 
which represents
the probability of 
an output sequence 
$\mathbf{y}_{1:h}^* \triangleq 
\{ \mathbf{s}_{k} \}_{k=1}^{h} \in \mathbb{R}^{d \times h}$
of 
$h \geq 1$ future 
$d$-dimensional states
of the vessel
given a new input sequence 
$\mathbf{x}_{1:\ell}^* \triangleq 
\{ \mathbf{s}_{k} \}_{k=1}^{\ell} \in \mathbb{R}^{d \times \ell}$
of $\ell \geq 1$ observed states,
and the available training data
$\mathcal{D}=\{\mathcal{X}^i,\mathcal{Y}^i,\Psi^i\}_{i=1}^{N}$ 
containing 
the two sets 
$\mathcal{X}^i = \{ \mathbf{x}_{1:\ell}^i \}_{i=1}^N$
and
$\mathcal{Y}^i =  \{ \mathbf{y}_{1:h}^i \}_{i=1}^N$
of training input and, respectively, output sequences.
Note that each sequence element $\mathbf{s}^i_k \triangleq \mathbf{s}^i(t_k) \in \mathbb{R}^d$ represents the vessel's position 
in latitude and longitude coordinates, 
taken from the available time-stamped AIS messages.
In many practical cases, it is also possible to exploit 
some additional information available from AIS data 
such as the vessel destination.
The destination port is an example of voyage related information provided by the AIS that may be relevant to anticipate a vessel trajectory.
We denote this (possibly available) input feature, 
which may be salient to predict the $i$-th output sequence,
by 
$\Psi^i = \{ \bm{\psi}^i \}_{i=1}^{N}$, 
where $\bm{\psi}^i \in \{0,1\}^v$ is the $v$-way categorical feature for each trajectory $i$ representing the class label of the specific motion pattern encoded into a one-hot vector of size $v$\cite{Capobianco2021}.
For example, 
$\bm{\psi}^i = \{0, 0, 1\}$ would mean that this particular vessel followed a trajectory that has been labeled $\#3$ based on three possible destinations.

\subsection{Modeling prediction uncertainty}

Uncertainty on the prediction estimates can be captured with
recently developed Bayesian deep learning tools,
which offer a practical framework for representing uncertainty in deep learning models \cite{Kendall2017,Bhatta2018,Xiao2019}.
In the context of supervised learning, 
two forms of uncertainty, i.e.,
\textit{aleatoric} and \textit{epistemic} uncertainty are considered,
where epistemic
is the reducible and aleatoric the irreducible part of uncertainty \cite{Kendall2017}.
Aleatoric (or data) uncertainty captures noise inherent in the observations,
whereas epistemic (or model) uncertainty accounts for uncertainty in the neural network model parameters \cite{Kendall2017}.
Epistemic uncertainty is a particular concern for neural networks
given their many free parameters, and can be large for
data that is significantly different from the training set.
Thus, for any real-world application of neural network uncertainty
estimation, it is critical that it be taken into account.
We follow a combined aleatoric-epistemic model \cite{Kendall2017} to capture both aleatoric and epistemic uncertainty in our prediction model.

Following a Bayesian framework \cite{Neal96} with prior distributions placed over the parameters of the NN, 
epistemic uncertainty can be captured by learning a distribution of NN models $p(F|\mathcal{D})$ representing the posterior distribution over the space of functions $\mathbf{y}_{1:h} = F(\mathbf{x}_{1:\ell})$ that are \emph{likely} to have generated our dataset $\mathcal{D}$.
The predictive probability \eqref{eq:predictive_distr} is then obtained by marginalizing over the implied posterior distribution of models, i.e.,
\begin{equation}
\label{eq:predictive_distr2}
p(\mathbf{y}_{1:h}^* | \mathbf{x}_{1:\ell}^*, \mathcal{D} ) = 
\int p(\mathbf{y}_{1:h}^* | \mathbf{x}_{1:\ell}^*, F )
p(F | \mathcal{D} ) d F .
\end{equation}
In our case, $F$ are RNN encoder-decoder models \cite{Capobianco2021}
assumed to be described by a finite set of parameters $\bm{\theta}$, such that
\begin{equation}
\label{eq:predictive_distr3}
p(\mathbf{y}_{1:h}^* | \mathbf{x}_{1:\ell}^*, \mathcal{D} ) = 
\int p(\mathbf{y}_{1:h}^* | \mathbf{x}_{1:\ell}^*, \bm{\theta} )
p(\bm{\theta} | \mathcal{D}) d \bm{\theta} .
\end{equation}
Since
$p(\bm{\theta} | \mathcal{D})$ cannot be obtained analytically,
it can be approximated by using variational inference with 
approximating distribution $q(\bm{\theta})$, which allows for efficient sampling. 
This results in the approximate predictive distribution 
\begin{equation}
\label{eq:predictive_distr4}
q(\mathbf{y}_{1:h}^* | \mathbf{x}_{1:\ell}^*, \mathcal{D} ) = 
\int p(\mathbf{y}_{1:h}^* | \mathbf{x}_{1:\ell}^*, \bm{\theta} )
q(\bm{\theta}) d \bm{\theta}, 
\end{equation}
which can be kept as close as possible to the original distribution 
by minimizing the Kullback-Leibler (KL) divergence
between $q(\bm{\theta})$ and the true posterior $p(\bm{\theta} | \mathcal{D})$ during the training stage.

Let us consider a generic NN architecture with $L$ transformation layers,
and denote by $\mathbf{W}_l$ the weight matrix of size $K_l \times K_{l-1}$ for each layer $l=1,\dots,L$. 
Then, following~\cite{Gal2016_nips}, by setting the set of weight matrices\footnote{All bias terms are omitted to simplify the notation.}
of the NN architecture 
as the set of parameters, i.e.,
$\bm{\theta} = \{\mathbf{W}_l\}_{l=1}^L$,
and
by using a Bernoulli approximating variational distribution, it is possible to relate variational inference in Bayesian NNs to the dropout mechanism.
In particular, the approximating distribution 
can be defined 
for each row $i$ of $\mathbf{W}_l$
as 

\begin{equation}
\label{eq:bernoulli}
q( \mathbf{w}_i ) = \gamma \, \mathcal{N}( \mathbf{w}_i; \mathbf{0}, \rho^2 \mathbf{I} ) 
+ ( 1-\gamma ) 
\mathcal{N}( \mathbf{w}_i; \mathbf{m}_i, \rho^2 \mathbf{I} ),
\end{equation}
where $\mathbf{m}_i$ is the vector of variational parameters, 
$\rho$ the standard deviation of the Gaussian prior distribution placed over $\mathbf{w}_i$,
and $\gamma$ the Bernoulli probability used for dropout.
Then, 
evaluating the model output 
$F_{\widehat {\bm{\theta}}}$
where sample $\widehat {\bm{\theta}} \sim q({\bm{\theta}})$
corresponds to performing dropout 
by randomly masking rows in the weight matrix $\mathbf{W}_l$ during the forward pass.
Predictions can then be obtained by performing 
MC dropout \cite{Gal2016} which consists of
executing dropout at test time and averaging results for all $M$ samples,
i.e.,
by approximating \eqref{eq:predictive_distr4} as 
\begin{equation}
\label{eq:predictive_distr5}
q(\mathbf{y}_{1:h}^* | \mathbf{x}_{1:\ell}^*, \mathcal{D} ) \approx 
\frac{1}{M} \sum_{j=1}^M 
p(\mathbf{y}_{1:h}^* | \mathbf{x}_{1:\ell}^*, \widehat {\bm{\theta}}_j ),
\end{equation}
with $\widehat {\bm{\theta}}_j \sim q({\bm{\theta}})$.
Note that in the case of 
MC dropout for RNNs, 
the same parameter realizations $\widehat {\bm{\theta}}_j$ 
are used for each time step of the input sequence \cite{Gal2016_nips}. 

As shown in \cite{Kendall2017}, aleatoric uncertainty can be modeled together with epistemic uncertainty 
by estimating the sufficient statistics of a given distribution describing the measurement noise of data. 
By fixing a Gaussian likelihood to model aleatoric uncertainty the predictive distribution of an
output sequence
$\hat{\mathbf{y}}_{1:h}$ for a given input sequence $\mathbf{x}_{1:\ell}$ 
can be approximated by a sequence of multivariate Gaussian distributions where each output is a Gaussian $\mathcal{N}(\mathbf{y}_{k};\hat{\mathbf{y}}_{k},\bm\Sigma_k)$ with predictive mean $\hat{\mathbf{y}}_{k}$ and covariance $\bm\Sigma_k, k=1,\dots,h$.

\subsection{Aleatoric and epistemic uncertainty}

The supervised learning task
\eqref{eq:predictive_distr}
can be recast as a sequence regression problem~\cite{Bishop96},
which aims at 
training a neural network model 
$F$ 
to predict, given an input sequence $\mathbf{x}_{1:\ell}$ of length $\ell$,
the 
predictive mean
$\hat{\mathbf{y}}_{1:h}$ 
and the predictive covariance $\bm\Sigma_{1:h}$, i.e.
\begin{equation}\label{eq:pred}
[ \hat{\mathbf{y}}_{1:h},\bm\Sigma_{1:h} ] = F_{\widehat {\bm{\theta}}} ( \mathbf{x}_{1:\ell} ) 
\end{equation} 
where $F$ is a BNN parameterized by model weights $\widehat {\bm{\theta}}$ drawn from the approximate dropout variational distribution $\widehat {\bm{\theta}} \sim q(\bm{\theta})$.
Note that a single network can be used to 
predict both
$\hat{\mathbf{y}}_{1:h}$ and $\bm\Sigma_{1:h}$.
By setting the distribution modeling aleatoric uncertainty as Gaussian, we induce a minimization objective of
the loss function $\mathcal{L}_{1:h} \triangleq \mathcal{L}( \mathbf{y}_{1:h} )$,  
here defined via the negative log-likelihood
\begin{eqnarray}\label{eq:neg_log_like}
\mathcal{L}_{1:h}
= \sum_{k=1}^h \frac{1}{2} (\mathbf{y}_{k} - \hat{\mathbf{y}}_{k})^T \bm\Sigma_{k}^{-1} (\mathbf{y}_{k} - \hat{\mathbf{y}}_{k}) 
+ \frac{1}{2} \operatorname{ln} |\bm\Sigma_{k}|
\end{eqnarray} \normalsize
which enables 
simultaneous training of both 
$\hat{\mathbf{y}}_{k}$ and $\bm\Sigma_{k}$
in an
end-to-end optimization process of the entire 
vessel prediction and uncertainty estimation framework.
Thus, for each state of the predicted sequence, the network outputs a Gaussian distribution parameterized by $\hat{\mathbf{y}}_{k}$ and $\bm\Sigma_{k}$
such that the negative log-likelihood in \eqref{eq:neg_log_like} of the ground truth vessel positions $\mathbf{y}_{k}, k=1,\dots,h$ over all training data is as small as possible. 

If the epistemic uncertainty can be estimated through MC dropout \cite{Gal2016,Kendall2017}
from $M$ samples  
$\{\hat{\mathbf{y}}_{1:h}^j,\bm\Sigma_{1:h}^j\}_{j=1}^M$, 
then the total predictive uncertainty for the output $\mathbf{y}_{k}$
can be approximated as follows:
\begin{eqnarray}\label{eq:var_tot} 
&&\hspace{-.7cm}
\overline{\bm\Sigma}_{k}
= \frac{1}{M} \sum_{j=1}^M \hat{\mathbf{y}}_{k}^j \hat{\mathbf{y}}_{k}^{j^{T}}
- \Bigg[ 
\frac{1}{M} \sum_{j=1}^M \hat{\mathbf{y}}_{k}^j 
\frac{1}{M} \sum_{j=1}^M \hat{\mathbf{y}}_{k}^{j^{T}} 
\Bigg]
\nonumber
\\
&&
+\frac{1}{M}
\sum_{j=1}^M \bm\Sigma_{k}^j,
\end{eqnarray} 
where the first two terms of the sum correspond to the epistemic uncertainty,
and the third corresponds to the aleatoric
uncertainty.
Finally, we average the uncertainty across time steps $k=1,\dots,h$
to obtain the uncertainty estimate of the complete output sequence.

\section{Learning to predict under uncertainty}
\label{sec:encdec}

We extend the encoder-decoder architecture 
based on recurrent networks with attention
proposed for
vessel trajectory prediction in~\cite{Forti2020,Capobianco2021}
by providing a practical quantification of the total (i.e., comprising both model and data: epistemic and aleatoric) predictive uncertainty following a Bayesian deep learning approach.

The encoder-decoder architecture consists of an bidirectional encoder RNN that reads a sequence of vessel positions one state at a time encoding the input information into a sequence of hidden states,
and a decoder RNN that generates an output sequence 
of future states step-by-step conditioned on a \emph{context} representation of the input sequence 
generated from the encoder's hidden states through an attention aggregation layer.
Moreover, we show that, if available, additional information on the intention of the vessel can be exploited to generate the decoder's outputs.

The RNN encoder-decoder architecture consists of an encoder model with $\theta_{E}$ parameters and a decoder model with $\theta_D$ parameters. Each model is implemented as a fully-trainable RNN containing different weight matrices. We define the Bernoulli variational distribution $q(\theta)$ over the union of all the weight matrices of our architecture, i.e., $\theta = \{\theta_E, \theta_D\}$. In the next sections we will see how the full model learns to predict future trajectories under uncertainty. 
\subsection{Variational LSTM}

LSTM networks are a special kind of RNN capable of capturing long temporal  patterns  in  sequential data. They work well on a large variety of problems~\cite{Capobianco2021,Forti2020,Graves2013}. 
A Bayesian view of RNNs has been proposed in~\cite{Gal2016_nips}, where the authors interpret LSTM as probabilistic models considering the network weights as variables trainable by suitable likelihood functions. 
This has been shown to be equivalent to implementing a 
novel variant of dropout for RNN models
to
approximate the posterior distribution over the weights with a mixture of Gaussians 
leading to 
a tractable optimization objective.

Following~\cite{Gal2016_nips}, we extend the deterministic LSTM architecture implementing dropout
on input or output, and recurrent connections,
with the same network units dropped at each time step.
In this paper, the LSTM architecture maps the input sequence $\mathbf{x}_{1}, \dots, \mathbf{x}_{\ell}$ into a sequence of cell activation and hidden states by applying the \textit{tied-weights} LSTM parameterization 
in \cite{Gal2016_nips}: 
\begin{equation}\label{eq:lstm_mat}
    \begin{bmatrix}
    \tilde{\mathbf{i}}_t\\
    \tilde{\mathbf{f}}_t\\
    \tilde{\mathbf{o}}_t\\
    \tilde{\mathbf{g}}_t
    \end{bmatrix} = \underbrace{\begin{bmatrix}
    \mathbf{W}_i & \mathbf{U}_i \\
    \mathbf{W}_f & \mathbf{U}_f \\
    \mathbf{W}_o & \mathbf{U}_o \\
    \mathbf{W}_g & \mathbf{U}_g
    \end{bmatrix}}_{\mathbf{W}_R} \begin{bmatrix}
    \mathbf{x}_t \\
    \mathbf{h}_{t-1}
    \end{bmatrix} = \begin{bmatrix}
    \mathbf{W}_i\mathbf{x}_t + \mathbf{U}_i\mathbf{h}_{t-1} \\
    \mathbf{W}_f\mathbf{x}_t + \mathbf{U}_f\mathbf{h}_{t-1} \\
    \mathbf{W}_o\mathbf{x}_t + \mathbf{U}_o\mathbf{h}_{t-1} \\
    \mathbf{W}_g\mathbf{x}_t + \mathbf{U}_g\mathbf{h}_{t-1}
    \end{bmatrix},
\end{equation}
where $\mathbf{W}_{i}, \mathbf{W}_f,\mathbf{W}_o, \mathbf{W}_g$ are all $p$-by-$d$ weight matrices, and $\mathbf{U}_{i}, \mathbf{U}_f,\mathbf{U}_o, \mathbf{U}_g$ have dimension $p$ by $p$, where $d$ is the number of input features and $p$ is the dimension of the (unidirectional) hidden state, i.e., $\mathbf{h}_{t-1} \in \mathbb{R}^p$. The \emph{input}, \emph{forget}, \emph{output}, and \emph{input modulation} gates are
$\mathbf{i}_t$, $\mathbf{f}_t$, $\mathbf{o}_t$ and $\mathbf{g}_t$, respectively, and can be computed as
\begin{IEEEeqnarray}{rClCrCl}\label{eq:ifog}
\mathbf{i}_t &=& \operatorname{sigm}(\tilde{\mathbf{i}}_t), &\qquad\qquad&
\mathbf{f}_t &=& \operatorname{sigm}(\tilde{\mathbf{f}}_t), \IEEEnonumber \\
\mathbf{o}_t &=& \operatorname{sigm}(\tilde{\mathbf{o}}_t), &\qquad\qquad&
\mathbf{g}_t &=& \tanh(\tilde{\mathbf{g}}_t).
\end{IEEEeqnarray}
%
Finally, the cell activation state and hidden state vectors, respectively $\mathbf{c}_t$ and $\mathbf{h}_t$, are: 
\begin{eqnarray}\label{eq:lstm}
\mathbf{c}_t &=& \mathbf{f}_t \odot \mathbf{c}_{t-1} + \mathbf{i}_t\odot\mathbf{g}_t \nonumber\\
\mathbf{h}_{t} &=& \mathbf{o}_t\odot \operatorname{tanh}(\mathbf{c}_{t}),
\end{eqnarray}
%
where $\odot$ denotes the Hadamard product. Note that the matrix $\mathbf{W}_R \in \mathbb{R}^{(4p)\times(d+p)}$ is a compact representation of all the weight matrices of the four LSTM gates.


In order to perform approximate variational inference over the weights, 
we may write the dropout variant of the parameterization in \eqref{eq:lstm_mat} as
\begin{equation}\label{eq:lstm_variational}
    \begin{bmatrix}
    \tilde{\mathbf{i}}_t\\
    \tilde{\mathbf{f}}_t\\
    \tilde{\mathbf{o}}_t\\
    \tilde{\mathbf{g}}_t
    \end{bmatrix} = 
    \mathbf{W}_R
    \begin{bmatrix}
    \mathbf{x}_t \odot \mathbf{d}_x\\
    \mathbf{h}_{t-1} \odot \mathbf{d}_h 
    \end{bmatrix} 
\end{equation}
with $\mathbf{d_x}$, $\mathbf{d_h}$ random masks repeated at all time steps. 
In the following sections, we will see how to apply the above Variational LSTM (\textit{VarLSTM}) model to our trajectory prediction task under uncertainty, using an encoder-decoder RNN architecture.

\subsection{Encoder Network}

The encoder network is designed as a bidirectional RNN~\cite{Schuster97} to capture and analyze the temporal patterns in vessel trajectory  both in the positive and negative time directions simultaneously. Following the encoder architecture used in~\cite{Capobianco2021}, we use two \textit{VarLSTM} to encode the input sequence using one model for each time direction. 

In our \textit{VarLSTM} encoder implementation, the dropout mask $\mathbf{d}_h$ is applied only to the recurrent connections, therefore it does not perturb the input vessel trajectory. In the end, the trainable parameters are $\bm{\theta}_E = \{\overrightarrow{\mathbf{W}}_R, \overleftarrow{\mathbf{W}}_R\}$, one weight matrix for each \textit{VarLSTM}.

The bidirectional \textit{VarLSTM} maps the input sequence 
into two different temporal representations:
the forward hidden sequence by iterating the
forward layer from
$t = 1$ to $\ell$, 
and the backward hidden sequence by iterating the backward layer from $t = \ell$ to $1$. 
In this way, the encoder network 
is able to learn long-term patterns in both temporal directions.
The output layer of the encoder is then updated  
into a compact hidden state representation 
obtained by 
concatenating the forward and backward hidden states, i.e.,
 $\overleftrightarrow{\mathbf{h}}_{t} = \overrightarrow{\mathbf{h}}_{t} \oplus  \overleftarrow{\mathbf{h}}_{t}$
computing the output vectors of the encoder layer $\overleftrightarrow{\mathbf{h}}_1,\dots,\overleftrightarrow{\mathbf{h}}_\ell$
for an input sequence of length $\ell$.
Each element $\overleftrightarrow{\mathbf{h}}_t$ encodes bidirectional spatio-temporal information of the input sequence extracted from the 
states of the vessel preceding and following the
$t$-th component of the sequence.

\subsection{Attention-based Decoder Network with Uncertainty} 
In the proposed architecture, the decoder RNN is trained to learn
the following conditional probability
\begin{equation}\label{eq:decoder2}
p(\mathbf{y}_{1}, \dots, \mathbf{y}_{h} | \mathbf{x}_{1}, \dots, \mathbf{x}_{\ell}) = \prod_{t=1}^{h} 
p(\mathbf{y}_{t}|\{\mathbf{y}_{1}, \dots, \mathbf{y}_{t-1}\},\mathbf{z}_t, \bm{\psi}),
\end{equation}
where each factor in \eqref{eq:decoder2} can be modeled by an RNN $D$, with trainable parameters $\bm{\theta}_{D}$, of the form
\begin{equation}\label{eq:decoder3}
    p(\mathbf{y}_{t}|\{\mathbf{y}_{1}, \dots, \mathbf{y}_{t-1}\},\mathbf{z}_t) = D(\hat{\mathbf{y}}_{t-1}, \mathbf{s}_{t-1}, \mathbf{z}_t;\bm{\theta}_{D}).
\end{equation}
The decoder~\eqref{eq:decoder3} generates the next kinematic state $\hat{\mathbf{y}}_{t}$
given the context vector $\mathbf{z}_t$,
the possibly available information on the high-level intention $\bm{\psi}$ of the vessel,
the decoder's hidden state $\mathbf{s}_{t-1}$,
and the previous predictions $\hat{\mathbf{y}}_{t-1}$,
which are fed back into the model in a recursive fashion 
as additional inputs to predict further into the future.
In addition, \eqref{eq:decoder3} is initialized by setting
\begin{equation}\label{eq:init_map}
    \mathbf{s}_{0} = g( \overrightarrow{\mathbf{h}_\ell}, \bm{\psi})
\end{equation}
to map the last encoder (forward) hidden state and the intention into the initial decoder hidden state $\mathbf{s}_{0}$. 
Note that, similar to~\cite{Capobianco2021}, this work partially addresses the multimodal nature of the prediction task with the use of the vessel's intention (i.e., destination) $\bm{\psi}$. However, different from~\cite{Capobianco2021} and as an additional measure to avoid overfitting, here the intention information is used only to initialize the decoder through \eqref{eq:init_map}, which in this case takes the form
 \begin{equation}\label{eq:init_map2}
    \bm{s}_0 = 
    \tanh(\mathbf{W}_{\psi}\bm{\eta})
\end{equation}
with
$\bm{\eta} = \overrightarrow{\mathbf{h}_\ell} \oplus \bm{\psi}$, 
$\mathbf{W}_{\psi}$ 
being the trainable parameters\footnote{\label{note_biases}Again, all biases are omitted for simplicity.}
of the neural network \eqref{eq:init_map}.

The attention mechanism~\cite{bahdanau2015} is adopted as an intermediate layer between the encoder and the decoder
to learn the relationship between the observed and the predicted kinematic states
while preserving the spatio-temporal structure of the input. We extend the attention module 
of \cite{bahdanau2015} implemented in \cite{Capobianco2021}
by applying a random dropout mask repeated at all time steps $\mathbf{d}_a$ to the input hidden features, previously computed by the encoder network. 
This is achieved by allowing the context representation to be a set of fixed-size vectors, or context set
$\mathbf{z}=\{\mathbf{z}_t\}_{t=1}^{\ell}$,
where
each context vector $\mathbf{z}_t$ in~\eqref{eq:decoder3}
can be computed as a weighted sum of the encoded input states, i.e.,
$
   \mathbf{z}_t =  \sum_{j=1}^\ell \alpha_{tj}\mathbf{h}_j \odot \mathbf{d}_a ,
$
where 
$
\alpha_{tj} = \operatorname{exp}(e_{tj}) 
/
\sum_{k=1}^\ell \operatorname{exp}(e_{tk})
$
represents the attention weight,
and $e_{tj} = a(\mathbf{s}_{t-1},\mathbf{h}_j \odot \mathbf{d}_a; \mathbf{W}_a)$
is a variational neural network with parameters 
$\mathbf{W}_a$
and dropout mask $\mathbf{d}_a$. 
The variational attention network is trained jointly with
the prediction model to quantify the level of matching between the inputs around position $j$ and the output at position $t$
based on the $j$-th encoded input state $\mathbf{h}_j$
and the decoder's state $\mathbf{s}_{t-1}$ (generating the $t$-th output).

In order to deal with uncertainty,
the decoder~\eqref{eq:decoder3} 
is implemented as a unidirectional \textit{VarLSTM} 
which generates the sequence of future predicted distributions by iterating $\forall t = 1, \dots, h$: 
\begin{eqnarray}
\bm{c}_t &=& \hat{\mathbf{y}}_{t-1} \oplus \mathbf{z}_t\\
\mathbf{s}_t &=& \operatorname{\textit{VarLSTM}}(\bm{c}_t, \mathbf{s}_{t-1}; \mathbf{W}_R)    
\label{eq:var_lstm}
\\
\hat{\mathbf{y}}_{t}
&=& \mathbf{W}_{\mu} \mathbf{s}_{t} 
\label{bnn_mean}
\\
 \left[ \log b_{11}, \log b_{22}, b_{21}\right]^T   &=& \mathbf{W}_{\Sigma} \mathbf{s}_{t} \label{bnn_variance}
\end{eqnarray}
where
$\hat{\mathbf{y}}_0 = \mathbf{x}_\ell$, 
$\mathbf{z}_t$ is the context vector computed through the attention mechanism, 
and $\mathbf{W}_{\mu}$, $\mathbf{W}_{\Sigma}$ are trainable parameters\footnoteref{note_biases} of a single neural network mapping the \textit{VarLSTM} output $\mathbf{s}_t$ 
to the parameters 
in~\eqref{bnn_mean}-\eqref{bnn_variance} 
used to estimate
the predictive uncertainty at time step $t$,
here modeled as a multivariate Gaussian distribution $\mathcal{N}(
\hat{\mathbf{y}}_{t}, \mathbf{\Sigma}_t)$. 
Note that, 
while the predictive mean 
$\hat{\mathbf{y}}_{t}$
can be directly obtained through~\eqref{bnn_mean},
to stabilize training and enforce positive-definite predictive covariance matrices,
the network~\eqref{bnn_variance} is trained to predict the elements of a lower triangular matrix with real and positive diagonal entries $\mathbf{B} = \begin{bmatrix}
b_{11} & 0 \\ b_{21} & b_{22}
\end{bmatrix}$,
such that
\begin{equation} \label{chol}
    \bm{\Sigma}_t = \mathbf{B}\mathbf{B}^T = \begin{bmatrix}
    \sigma_{t,x}^2 & \sigma_{t,x}\sigma_{t,y}\rho_t  \\
    \sigma_{t,x}\sigma_{t,y}\rho_t & \sigma_{t,y}^2
  \end{bmatrix}
\end{equation}
is guaranteed to be positive definite using the Cholesky decomposition~\eqref{chol}. 
Note that $\sigma_{t,x}$ and $\sigma_{t,y}$ in~\eqref{chol} denote the components of the predictive variance $\sigma_{t}$ along the $x$ and, respectively, $y$ direction.
Notice also that in the proposed \textit{VarLSTM} decoder implementation, the $\mathbf{d}_h$ dropout  mask  is  applied only on the recurrent connections.

This end-to-end solution is trained by the stochastic gradient descent algorithm in order to learn an optimal function approximation \eqref{eq:pred}.
A decoder \eqref{eq:decoder3} with such a recursive structure 
offers the advantage of being able to handle 
sequences of arbitrary length. 
The model consists of a set of trainable parameters $\bm{\theta}_{D} = \{\mathbf{W}_{\psi},\mathbf{W}_R,\mathbf{W}_a, \mathbf{W_\mu}, \mathbf{W}_\Sigma\}$ respectively used to model the decoder initialization \eqref{eq:init_map2}, the unidirectional \textit{VarLSTM} \eqref{eq:var_lstm}, the attention mechanism, and the output distribution given by \eqref{bnn_mean}-\eqref{bnn_variance}.

\begin{figure}[t!]
	\centering%
	\includegraphics[width=\columnwidth]{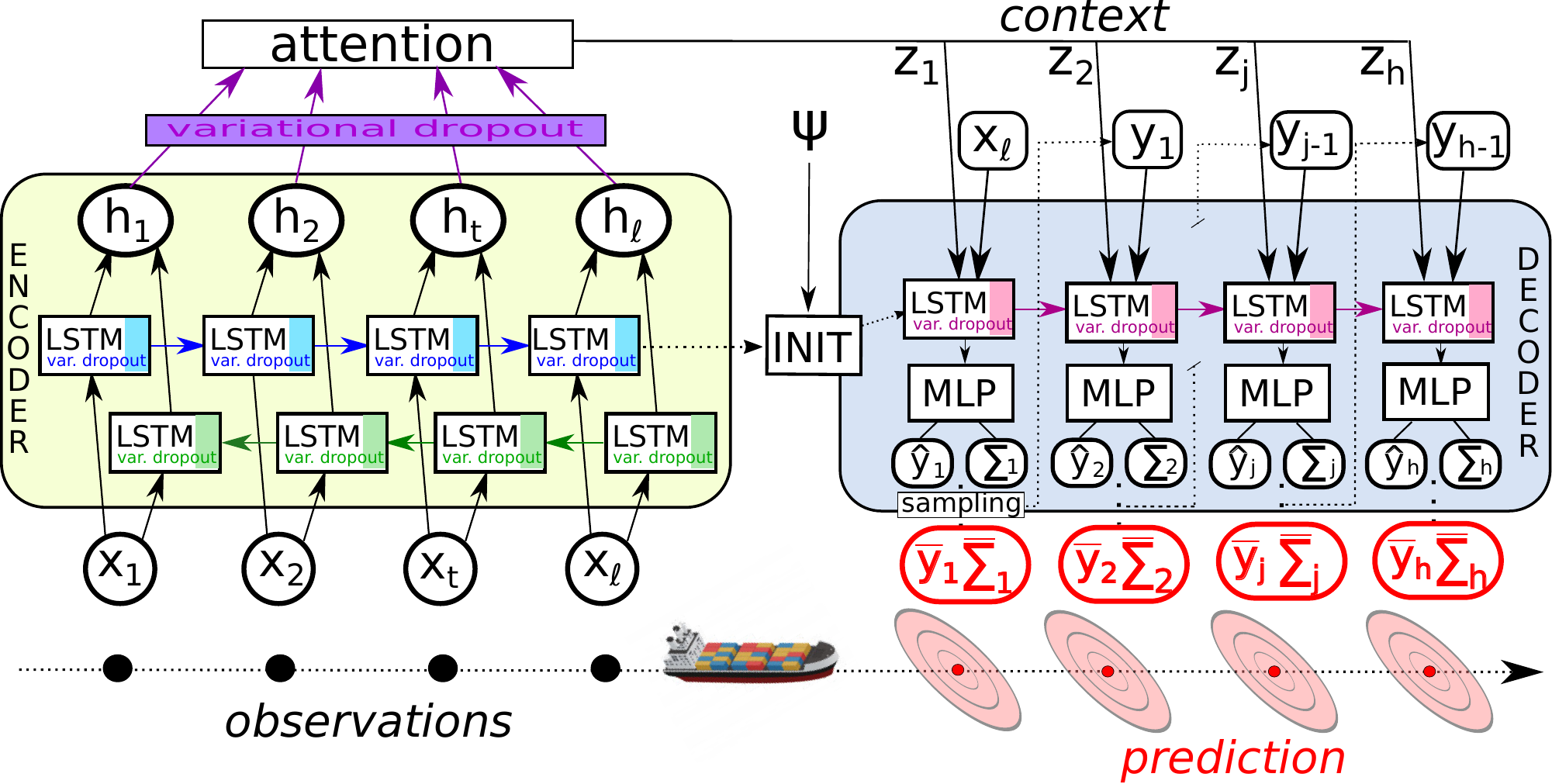}%
	\caption{Diagram of the proposed attention-based \textit{VarLSTM} encoder-decoder architecture for vessel trajectory prediction with uncertainty estimation. 
    The encoder is a bidirectional \textit{VarLSTM} that maps the input sequence into a sequential hidden state.
    Then, the encoded information, randomly masked through variational dropout, is passed to an intermediate attention module to generate a context representation of the input.
    This intermediate representation is fed to the decoder network, implemented as a unidirectional \textit{VarLSTM}, that generates the sequence of output distributions $\mathcal{N}(\hat{\mathbf{y}}_{t}, \mathbf{\Sigma}_t), t = 1, \dots, h$. Finally, the sequence of total predictive distributions $\mathcal{N}(\overline{\mathbf{y}}_{t}, \overline{\mathbf{\Sigma}}_t)$ is obtained by combining aleatoric and epistemic uncertainty, the latter estimated through MC dropout.
    Note that the decoder network is initialized using a regularization method on the intention information.}
	\label{fig:arch}
\end{figure}

 \subsection{Intention Regularization}\label{sec:regular} 

Despite the proposed decoder initialization~\eqref{eq:init_map2}, the encoder-decoder architecture may overfit the intention information, and predict erroneous future maneuvers of the vessel. 
This is due to the fact that the model tends to generate predictions that are mainly based on the high level intention rather than on the information encoded in the past trajectory. 
To avoid this behavior, and inspired by the dropout mechanism~\cite{Srivastava2014}, 
we apply a regularization technique based on dropout noise \cite{Wager2013}
by randomly masking the intention information at train time 
in order to prevent complex co-adaptations between the encoded past trajectory and the high-level information. 
 
In particular, 
we apply a random dropout noise to the intention at each iteration of the training procedure
by feeding
the intention information $\bm{\psi}$ into \eqref{eq:init_map2} with some probability $q$ (a predefined hyperparameter), or 
setting 
it to zero otherwise.
Thus, the decoder initialization~\eqref{eq:init_map2} takes the following form
\begin{equation}
   \bm{s}_0 = \tanh(   \mathbf{W}_{\psi}(\bm{\eta} \odot \mathbf{d}_{\eta} )), \\
\end{equation}
where 
$\mathbf{d}_{\eta} = \frac{p+v}{p} \bm{\nu}$,
and $\bm{\nu} \in \{0,1\}^{p+v}$ is a random binary mask on the input information such that $\nu_i=1$, $i=1,\dots,p$,
and $\nu_i= \nu \sim \operatorname{Bern}(1-\gamma)$, $i=p+1,\dots,p+v$. The random variable $\nu$
is Bernoulli with parameter $1-\gamma$, i.e., it takes the value $0$ with probability $\gamma$ for each training sample of the intention components of $\bm{\eta}$, otherwise it is $1$.
In other words, this random mask is applied to $\bm{\eta}$ in order to regularize via dropout noise only the intention information $\bm{\psi} \in \{0,1\}^v$, while leaving the encoded feature $\overrightarrow{\mathbf{h}_\ell} \in \mathbb{R}^p$ untouched. 
%
Then, a scaling factor 
$\frac{p+v}{p}$
is applied to the random mask at train time, while leaving the forward pass at test time unchanged.
Note that this pre-scaling performed at train time, commonly referred to as the \textit{inverted dropout} implementation, does not require any changes to the network to compensate for the absence of information during test time, as usually done in traditional dropout \cite{Srivastava2014}.

This additional regularization of the intention information 
can be viewed as a Multimodal Dropout method \cite{Neverova2016}, in which the input features belonging to the same
group (or modality) are either all dropped out or all preserved to avoid false co-adaptations between different groups of input information, and to handle missing data in one of the groups at test time.
In our case, this has been shown to improve the prediction performance by preventing co-adaptations between the encoded past observations and the possibly available intention information. 
The improvements in performance with respect to a previous version of the labeled architecture \cite{fusion2021} are provided in appendix \ref{appendix} for a varying probability $\gamma$.


\begin{figure}
    \centering%
    \includegraphics[trim=52 10 52 10,clip,width=.95\columnwidth]{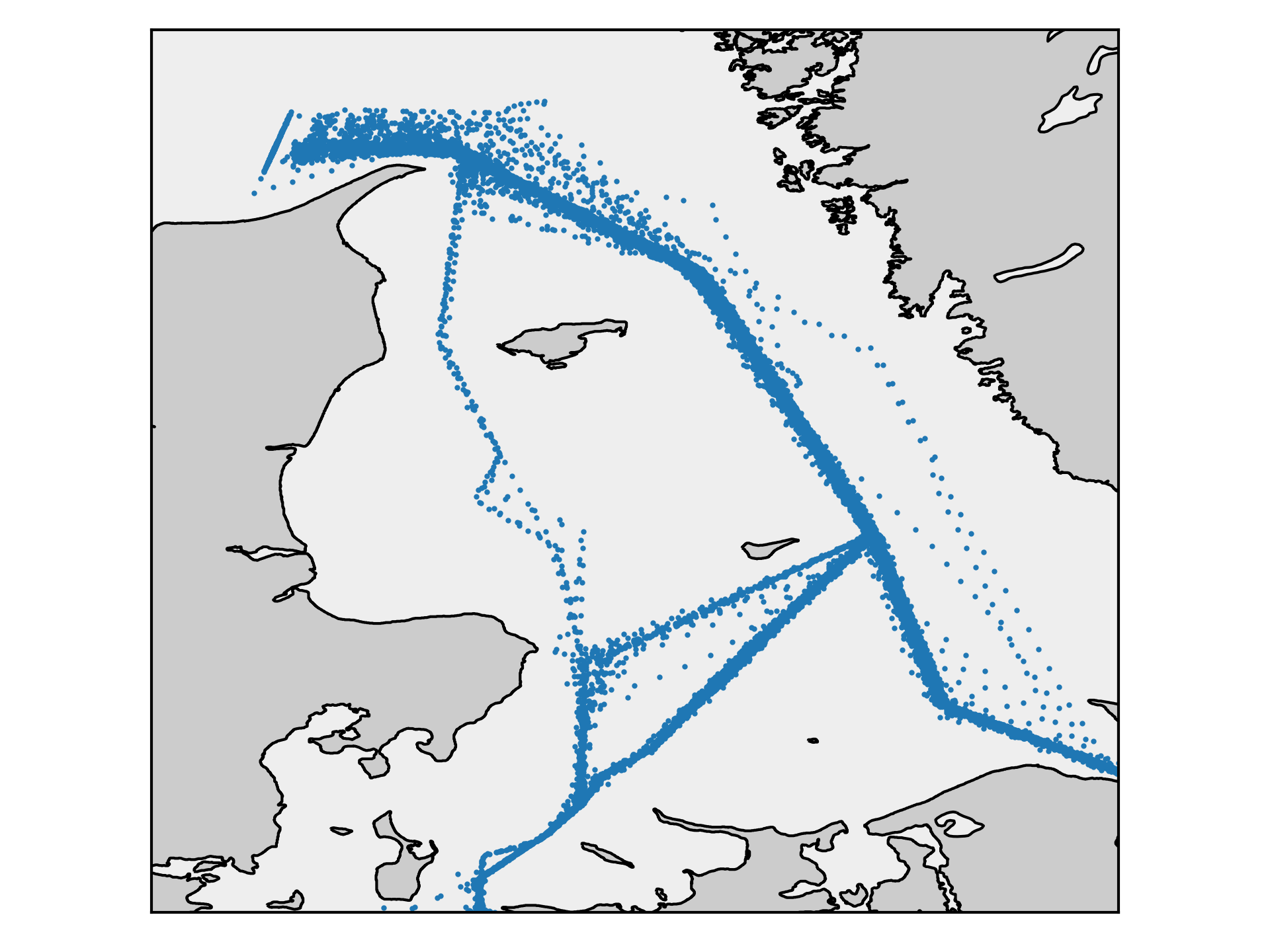}%
    \caption{Complete dataset of AIS positions used in the experiments; the training and testing sets are subsets of the dataset showed here. Each dot in the image corresponds to a ship's position. Contains data from the Danish Maritime Authority that is used in accordance with the conditions for the use of Danish public data~\cite{DMA}.}%
    \label{fig:training_dataset}%
\end{figure}

\begin{figure}
    \centering%
    \includegraphics[trim=137 10 137 10,clip,width=.95\columnwidth]{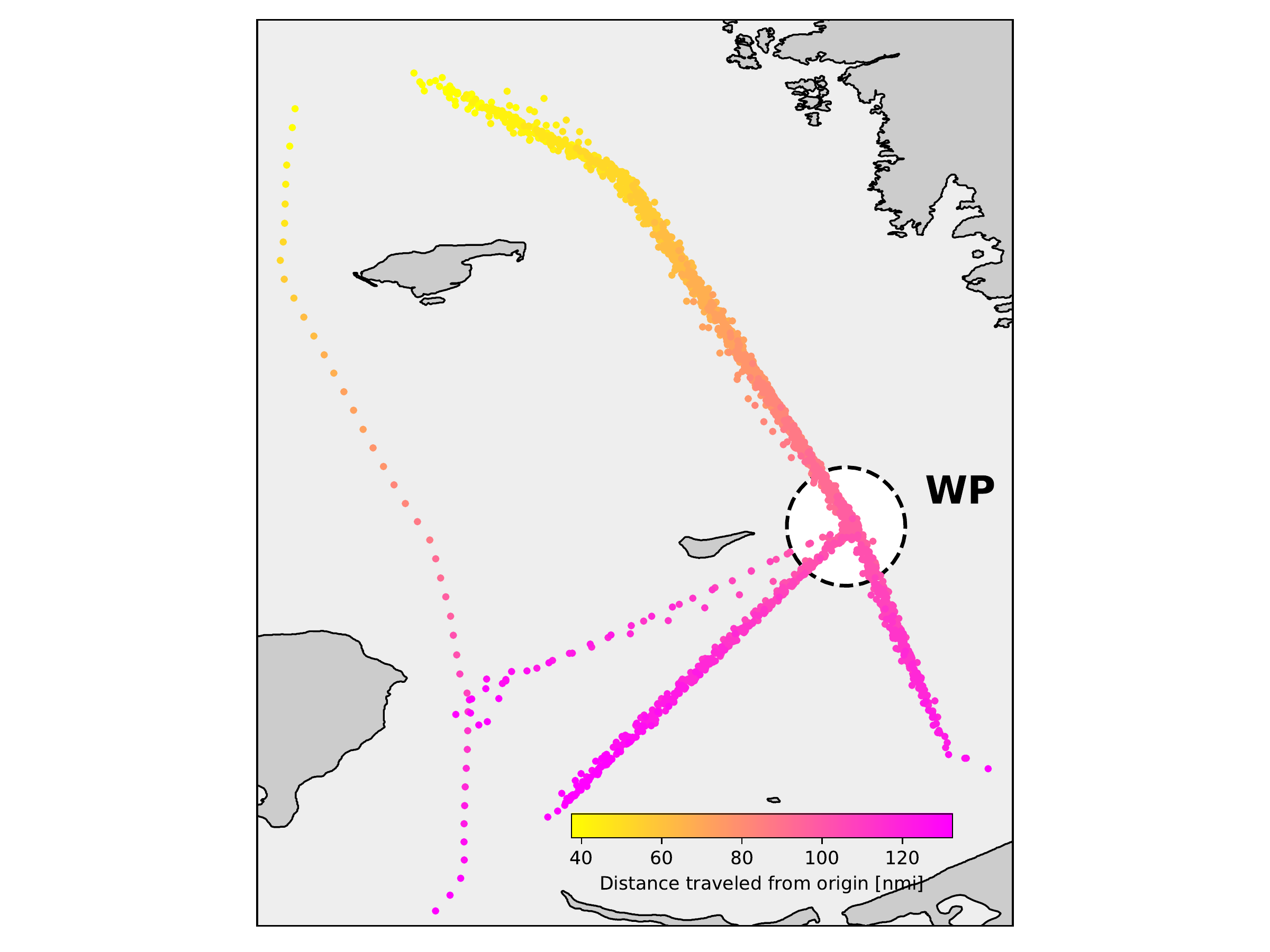}%
    \caption{Test dataset used for the evaluation of performance metrics. Each dot in the image corresponds to a ship's position and is colored based on its distance from a fixed point, located in the upper left corner of the map. Contains data from the Danish Maritime Authority that is used in accordance with the conditions for the use of Danish public data~\cite{DMA}.}%
    \label{fig:compare_dist_data}%
\end{figure}

\section{Experiments}\label{sec:experiments}

In this section we describe the experiments carried out to evaluate the proposed encoder-decoder architecture for trajectory prediction with uncertainty. 
\subsection{Dataset benchmark}
We used an AIS dataset extracted 
from the historical data made freely available by the Danish Maritime Authority (DMA)~\cite{DMA},
comprising $394$ trajectories of \textit{tanker} vessels belonging to two specific motion patterns of interest in the period ranging from January to February 2020. The complete dataset used in this paper is illustrated in Fig.~\ref{fig:training_dataset}, and a detailed description of the dataset preparation method can be found in~\cite{Capobianco2021}.
Paths falling into the same maritime pattern are described as sequences of positions
in planar coordinates assigned through the Universal Transverse Mercator (UTM) projection (zone 32V),
and correlated by voyage-related attributes including departure and destination areas.

In order to provide regular input sequences to 
train the model in a supervised fashion,
we applied a temporal interpolation to each trajectory using a fixed sampling time of 
$\Delta = 15$ minutes to resample the data,
and data segmentation by using the sliding window approach
to produce fixed-length input and output sequences 
of length $\ell=h=12$ vessel positions (i.e., $3$ hours). 

\subsection{Models}
We propose an encoder-decoder architecture with attention mechanism composed of 
a BiLSTM encoder layer with $64$ hidden units, 
and an LSTM decoder layer with $64$ hidden units.
For Bayesian modeling of the epistemic uncertainty, we used  MC dropout \cite{Gal2016} with $M=100$ samples, 
and dropout rate applied to recurrent connection $p = 0.05$ in both encoder and decoder layers.
The model was trained by applying AdamW optimizer~\cite{Loshchilov19}
with a learning rate of $0.0001$ and weight decay of $0.0001$ to minimize the mean absolute error loss function,
an early-stopping rule
with $3000$-epoch patience,
and a mini-batch size of $200$ samples.

In \cite{Ristic08} it is shown how 
better performance in terms of prediction accuracy can be achieved
by labeling the input data based on a high-level pattern information $\bm{\psi}$,
such as the vessel's intended destination.
In this regard, we compare the following two different prediction methods
based on labeled or unlabeled trajectories, where
in the \textit{labeled} case the neural model is trained to exploit also $\bm{\psi}$.  

\begin{figure*}
    \centering%
    \subfloat[][]{%
       \includegraphics[trim=5 10 0 10,clip,width=.95\columnwidth]{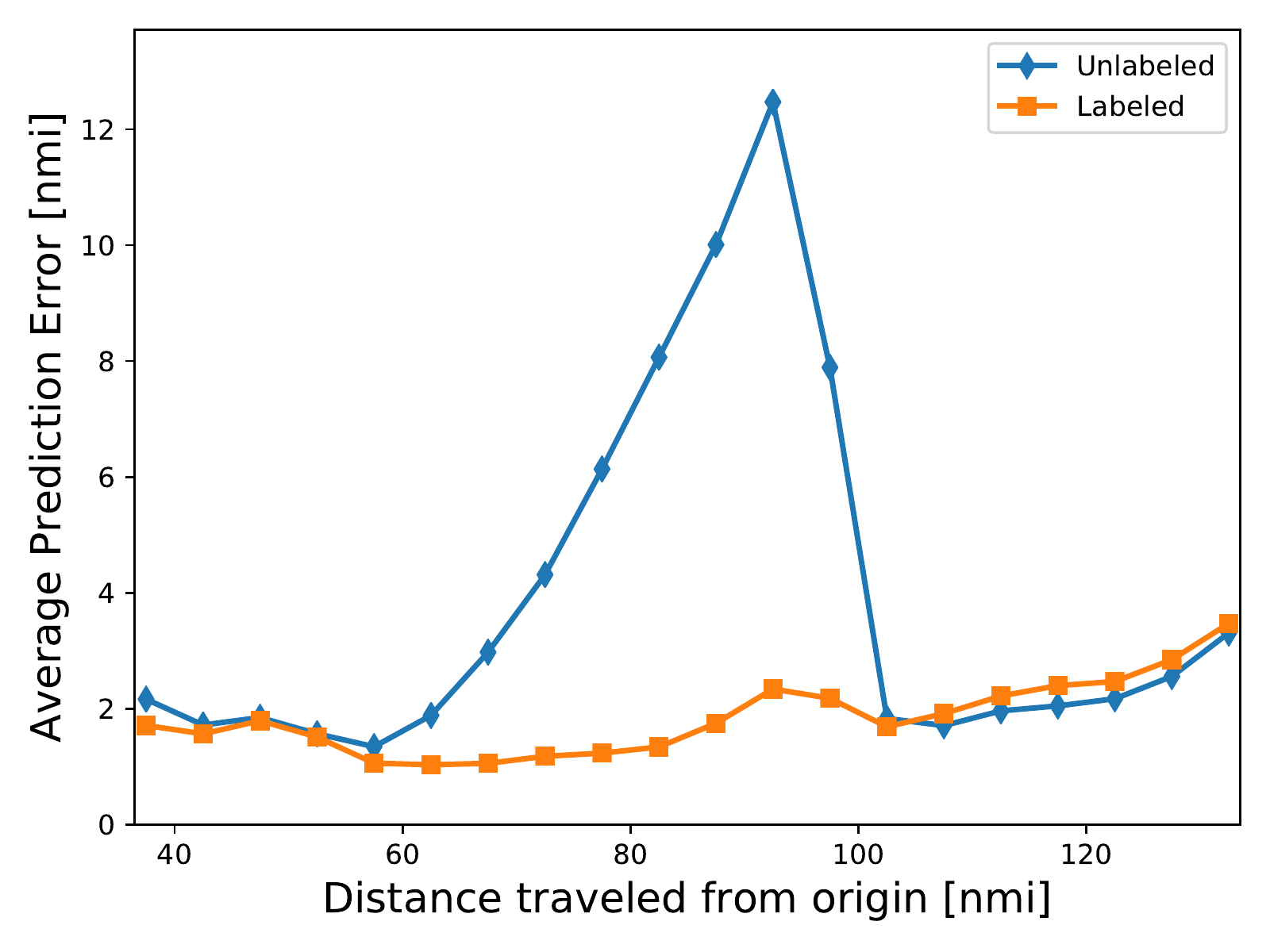}%
        \label{fig:compare_dist_perf}
        }%
    \hfil %
    \subfloat[][]{%
       \includegraphics[trim=5 10 0 10,clip,width=.95\columnwidth]{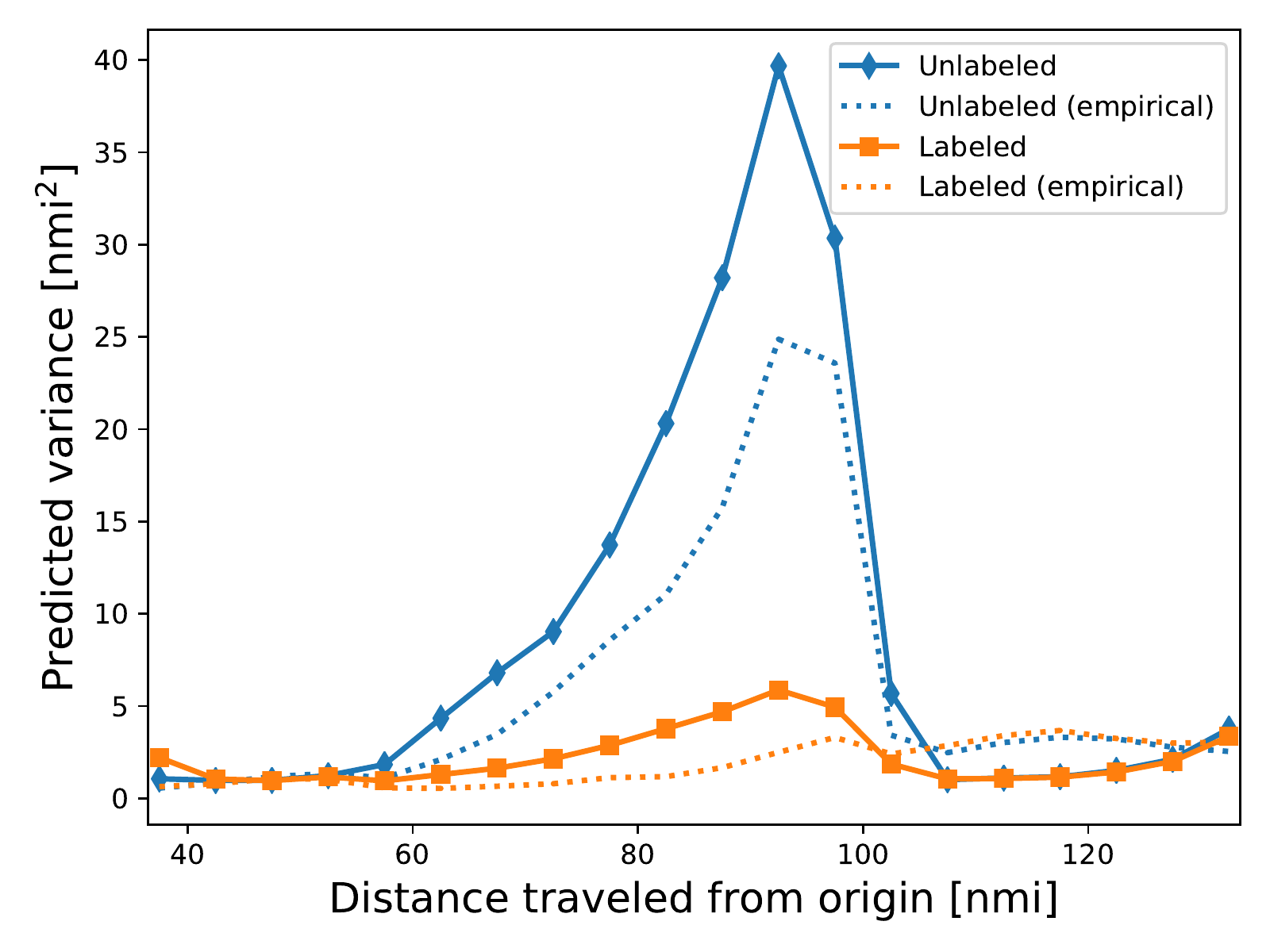}%
        \label{fig:compare_cov_perf}
        }%
    \caption{APE~\protect\subref{fig:compare_dist_perf} and predicted total variance~\protect\subref{fig:compare_cov_perf}. Both are computed at the $h$-th prediction sample (i.e., prediction horizon 3 hours) and are expressed as a function of the vessel distance from a fixed point, located in the upper left corner of Fig.~\ref{fig:compare_dist_data}. Panel~\protect\subref{fig:compare_cov_perf} shows the square root of the determinant of the covariance matrix produced by the network, which is also known as generalized variance and is proportional to the area of the prediction uncertainty ellipse.}%
    \label{fig:performance_comparison}%
\end{figure*}

\begin{itemize}
    \item \textit{Unlabeled (U)}: we train the predictive model and perform prediction using unlabeled 
    data, i.e. 
    using only
    low-level context representation encoded from 
    a sequence of 
    past observations,
    without any
    high-level
    information about the motion pattern. 
    \item \textit{Labeled (L)}: we train the predictive model and perform prediction using labeled 
    data, i.e.
    using
    low-level context representation encoded from 
    a sequence of 
    past observations, 
    as well as
    additional inputs $\bm{\psi}$
    about high-level intention behavior of the vessel. 
    This model includes the intention regularization mechanism with dropout probability $\gamma=0.3$, shown to be the highest-performing model in appendix \ref{appendix}.
\end{itemize}

\subsection{Results}

In this section we demonstrate the effectiveness of the proposed 
encoder-decoder \textit{VarLSTM} 
in learning trajectory predictions with quantified uncertainty for the maritime domain using labeled and unlabeled data.
From the complete dataset in Fig.~\ref{fig:training_dataset}, we isolated a set of trajectories to be used as test set. The selected test trajectories are illustrated in Fig.~\ref{fig:compare_dist_data}, with a color that changes with the distance traveled from a fixed origin point, located in the upper left corner of the figure. 
Figure~\ref{fig:compare_dist_perf} shows the Average Prediction Error (APE) computed as the average Euclidean distance between the predicted position and the ground truth at the $h$-th prediction sample (i.e., prediction horizon 3 hours)
as a function of the vessel's 
traveled distance
from the fixed origin point. 
This representation allows mapping all prediction errors on a common axis and identifying 
regions where the prediction error is high.

Figure~\ref{fig:compare_dist_perf} shows that 
the two models achieve comparable prediction errors,
apart from
a specific waypoint area $WP$ (distance between $80$ and $100$ nmi), highlighted in Fig.~\ref{fig:compare_dist_data},
where the APE obtained with the unlabeled model is much higher than that of the labeled model. 
This is a crossroad area where vessels can take three different paths towards the same destination,
which makes it challenging for the prediction system to anticipate which direction the vessel will follow after the crossroad. 
The plot proves how
high-level information is key to improve prediction performance corresponding to deviation areas (e.g., area $WP$ in Fig. \ref{fig:compare_dist_data}).
The major difference between unlabeled and labeled predictions is that unlabeled 
use past observed positions to generate future states, while labeled 
use additional high-level pattern information, i.e., 
the ship's intended destination.
In the area $WP$ (Fig.~\ref{fig:compare_dist_data}), 
the unavailability of any prior information 
puts the unlabeled model at a disadvantage
for deciding which future path to be followed by the vessel after a crossroad.
Instead, the labeled model has more information available, and can generate 
more realistic future trajectories by exploiting the information related to the pattern descriptor.

In Fig.~\ref{fig:compare_cov_perf} we show the performance in terms
of predictive uncertainty estimation by using the notion 
of \textit{generalized variance}, defined as the determinant of the covariance matrix $\overline{\bm\Sigma}_{k}$ in \eqref{eq:var_tot} produced by the unlabeled and labeled models; more precisely, in Fig.~\ref{fig:compare_cov_perf}, the square root of the determinant of the prediction covariance matrix (averaged over all the trajectories) is plotted, which is proportional to the area of the uncertainty ellipse.

\begin{figure*}
\centering %
\subfloat[]{%
    \label{fig:predictions_unlabeled_1}
    \includegraphics[trim=50 10 50 10, clip, width=.3\textwidth]{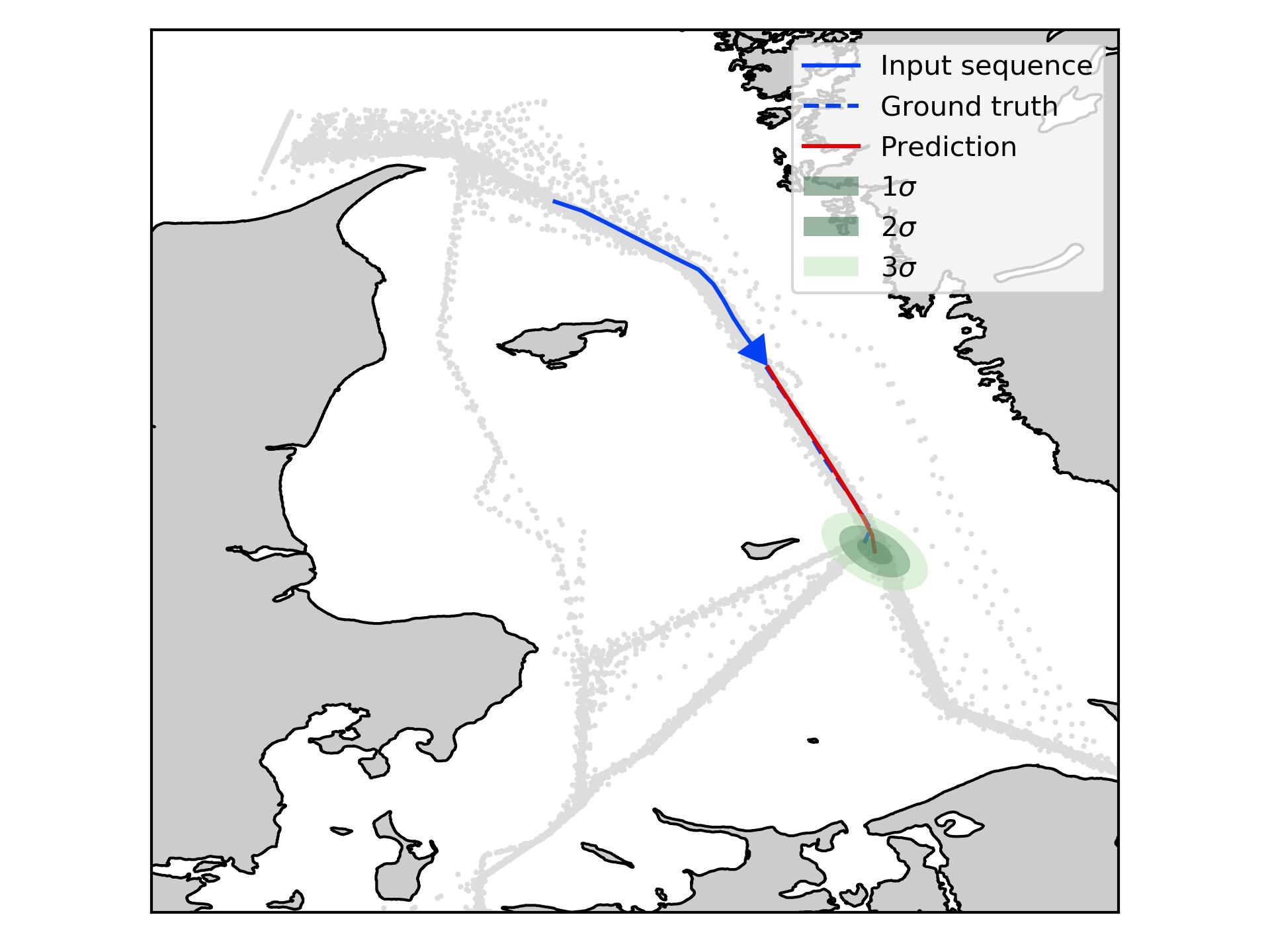}%
    } \hfill
\subfloat[]{%
    \label{fig:predictions_unlabeled_2}
    \includegraphics[trim=50 10 50 10, clip, width=.3\textwidth]{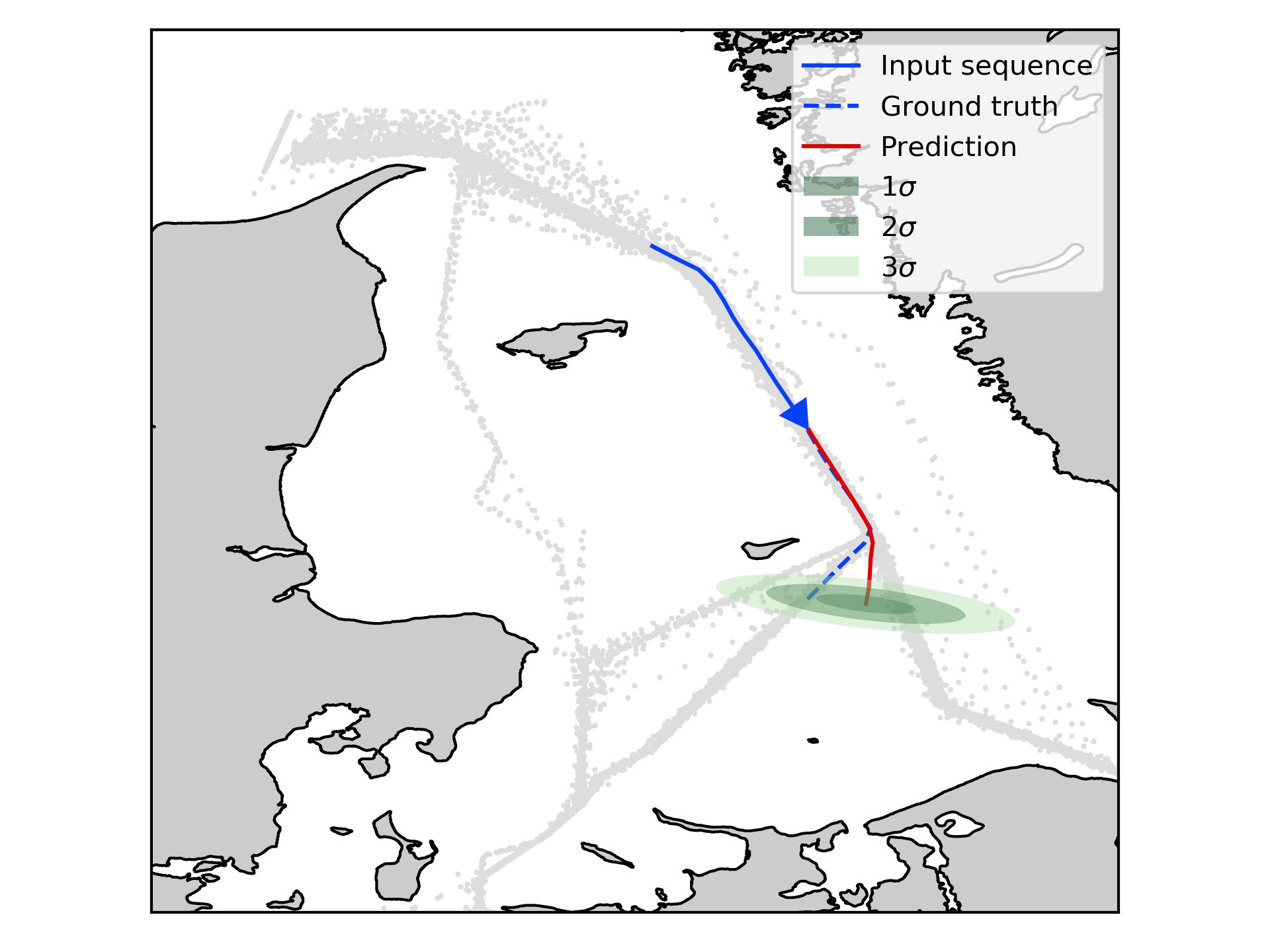}%
    } \hfill
\subfloat[]{%
    \label{fig:predictions_unlabeled_3}
    \includegraphics[trim=50 10 50 10, clip, width=.3\textwidth]{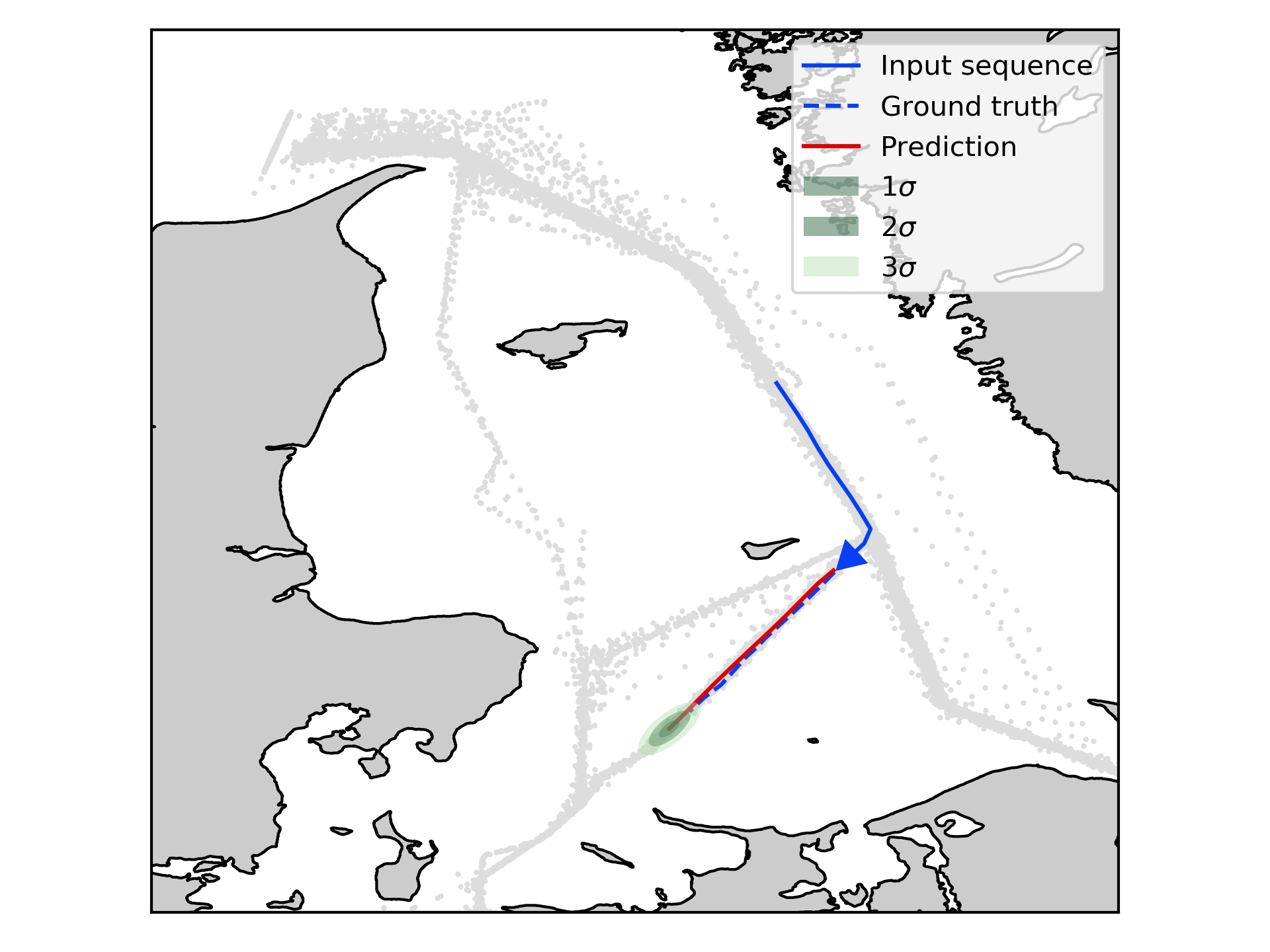}%
    } \\
\subfloat[]{%
    \label{fig:predictions_labeled_1}
    \includegraphics[trim=50 10 50 10, clip, width=.3\textwidth]{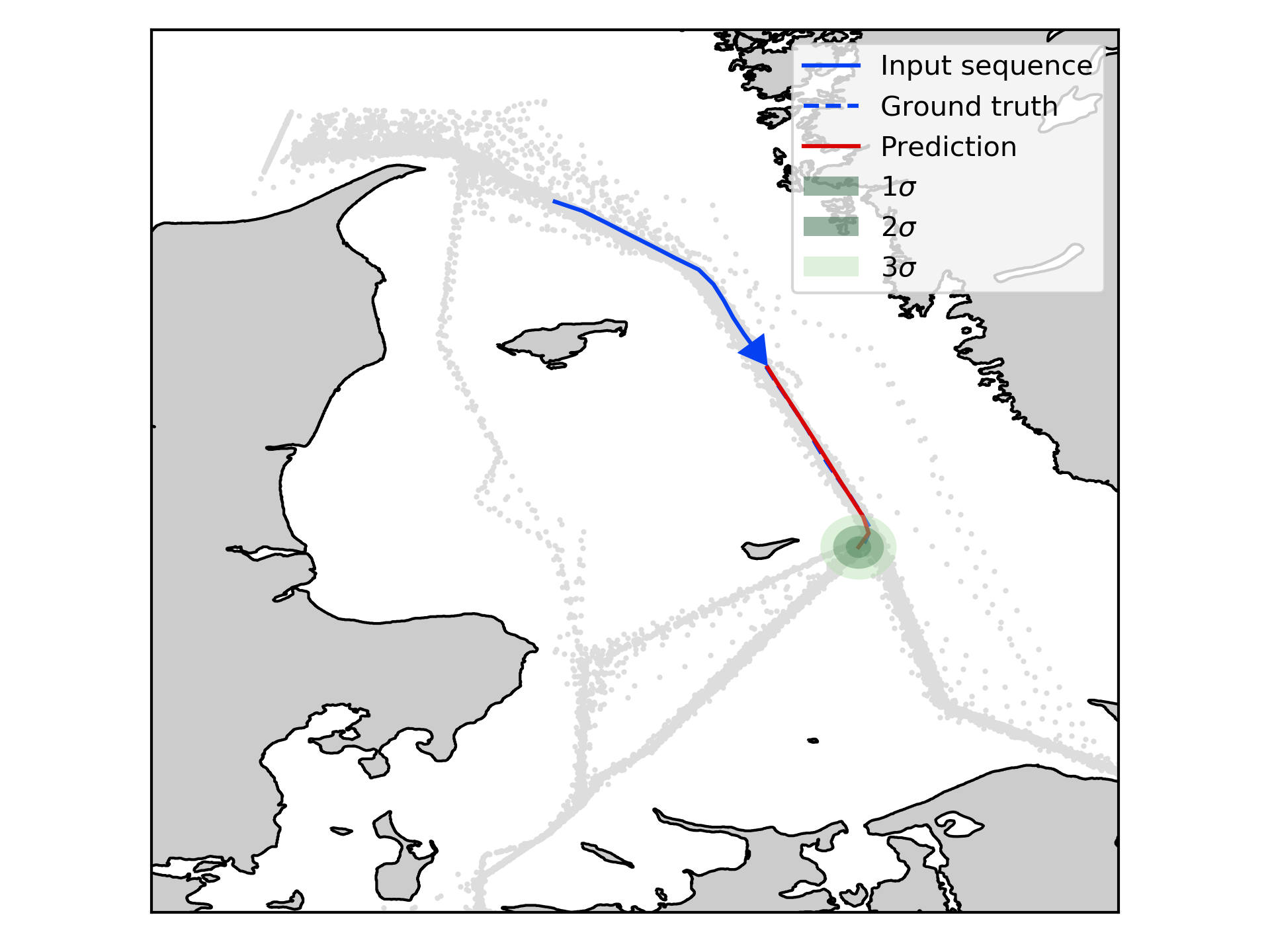}%
    } \hfill
\subfloat[]{%
    \label{fig:predictions_labeled_2}
    \includegraphics[trim=50 10 50 10, clip, width=.3\textwidth]{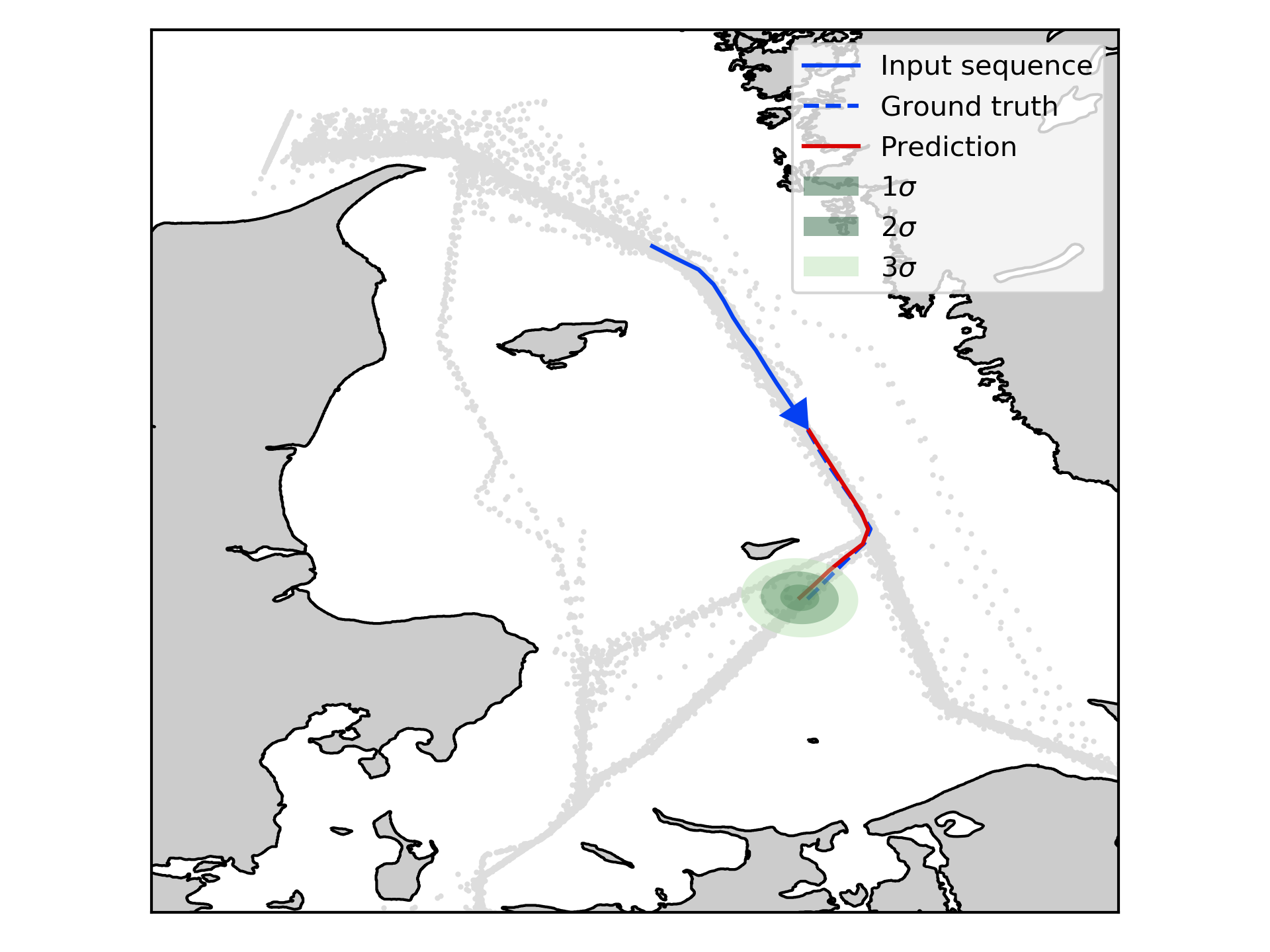}%
    } \hfill
\subfloat[]{%
    \label{fig:predictions_labeled_3}
    \includegraphics[trim=50 10 50 10, clip, width=.3\textwidth]{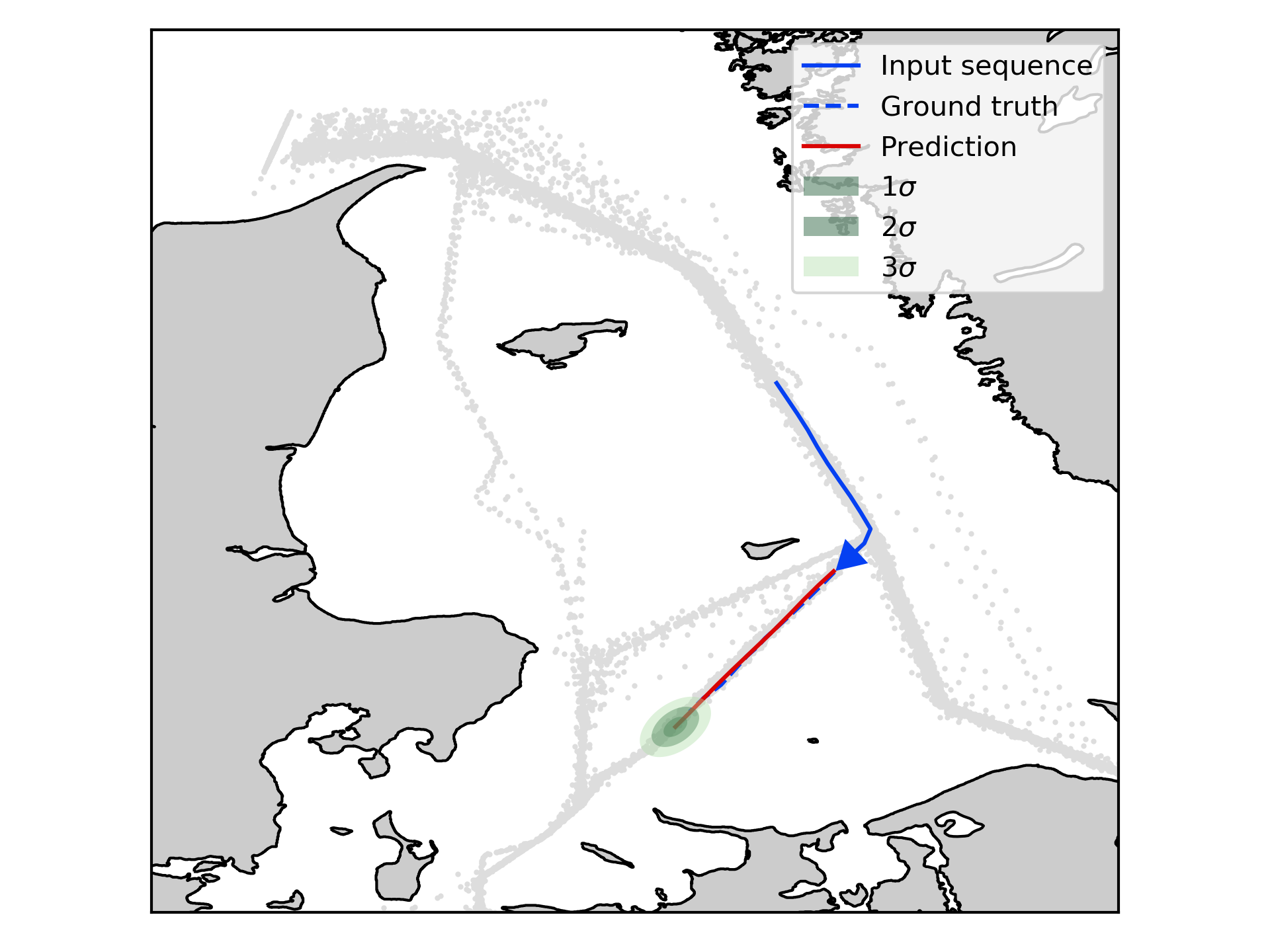}%
    }
\caption{Predictions computed with unlabeled models~\protect\subref{fig:predictions_unlabeled_1}--\protect\subref{fig:predictions_unlabeled_3} compared to labeled models~\protect\subref{fig:predictions_labeled_1}--\protect\subref{fig:predictions_labeled_3}. Contains data from the Danish Maritime Authority that is used in accordance with the conditions for the use of Danish public data~\cite{DMA}.}
\label{fig:predictions}
\end{figure*}

Figure~\ref{fig:predictions} shows the results obtained by the proposed model on a specific trajectory, which intersects the area $WP$ (Fig.~\ref{fig:compare_dist_data}) 
at three time steps $k_i$, $i=1,2,3$.
We perform the predictions considering the sliding window approach, so the input sequence is fed in each neural model to predict the next output sequence. 
We can see that the predictions using the 
unlabeled model 
shown in Figs.~\ref{fig:predictions_unlabeled_1}--\ref{fig:predictions_unlabeled_3}
detect the correct direction to be followed only after the critical point of a crossroad at time step $k_3$,
while, conversely, the ones using the
labeled model shown in Figs.~\ref{fig:predictions_labeled_1}--\ref{fig:predictions_labeled_3} are able to 
anticipate the direction to be followed by the vessel after the crossroad many steps ahead.
Another important point to be made regarding the predictive uncertainty modeling is
how the level of uncertainty quantified by the unlabeled model is larger and includes many possible directions when the path to be taken is still undetermined (e.g., in correspondence to a crossroad such as in Fig.~\ref{fig:predictions_unlabeled_2}). The dataset used to train the network (represented in Fig. \ref{fig:training_dataset}) contains multiple paths that can be taken by a vessel after the waypoint; for this reason, the unlabeled model computes a predicted trajectory that is the average mode among multiple possible paths, even if the unimodal prediction is not necessarily a valid future behaviour, and this is especially apparent in Fig.~\ref{fig:predictions_unlabeled_2}.
In contrast, the predictive uncertainty diminishes when the decision of the predictor based on the training data is less complicated due to the direct path of the vessel as shown in Fig.~\ref{fig:predictions_unlabeled_3}.

In the labeled case, due to the additional high-level information available, the predictions are shown to be more accurate, especially when the vessel is before the crossroad, as showed in Fig.~\ref{fig:predictions_labeled_1} and Fig.~\ref{fig:predictions_labeled_2}.
The labeled model can compute more accurate predictions, as some of the possible future paths (specifically, after the waypoint) can be excluded, precisely on the basis of the additional high-level information. This can be easily noted with a comparative inspection of Figures~\ref{fig:predictions_unlabeled_2} and~\ref{fig:predictions_labeled_2}, where the uncertainty of the unlabeled model, contrary to the labeled one, has to account also for the additional South-East path. It should be noted that the intention information is most useful when it can \textit{reduce} the multimodal nature of the task to a unimodal prediction task. For instance, in Fig.~\ref{fig:predictions_labeled_2}, the high-level information allows excluding the South-East path, but cannot help further in deciding \textit{exactly} which one of the two South-West paths the vessel is going to follow. Still, the labeled model achieves better prediction performance, as the additional information allows going from a higher modality prediction task to a lower modality one.

\section{CONCLUSION}
\label{sec:discussion}

In this paper, we proposed 
an attention-based recurrent encoder-decoder architecture to address the problem of 
trajectory prediction with uncertainty quantification applied to a maritime domain case study.
The predictive uncertainty is estimated through Bayesian learning by combining both aleatoric and epistemic uncertainty, with the latter modeled via Monte Carlo dropout.
Experimental results show that the proposed architecture is able to learn maritime sequential patterns from historical AIS data, and successfully predict 
future vessel trajectories with a reliable quantification of the predictive uncertainty.
Two models are compared and show how prediction performance can be improved by exploiting high-level intention behavior of vessels (e.g., their intended destination) when available. Future lines of research on this topic include the investigation of multimodal prediction techniques in combination with high-level intention modeling to further improve the prediction performance when the intention information alone is not sufficient to fully account for the multimodality of the prediction task.

\appendices

\section{Intention Regularization Performance}\label{appendix}

In this appendix we investigate how the prediction model can benefit from the intention regularization technique proposed in Section \ref{sec:regular} for the novel version of the labeled architecture (Lv2), by comparing the results with those obtained with the previous version of the labeled architecture (Lv1) presented in \cite{fusion2021}, in which the intention information is injected for each time step during the decoding phase.
For this comparison, we used the complete AIS dataset from \cite{Capobianco2021} shown in  Fig.~\ref{fig:training_dataset}. 
In our experiment we split the original dataset composed of \num{394} full trajectories into \num{284} trajectories for training, \num{32} for validation, and \num{78} for testing. 
The windowing procedure proposed in \cite{Capobianco2021} produces \num{8574} input/output sequences of length $\ell=h=12$ for training, \num{1054} sequences for validation, 
and \num{2379} sequences for the testing phase. 
In the evaluation of our experiment, 
we use 
the APE 
for different horizons (i.e., 1, 2, and 3 hours),
and the Average Displacement Error (ADE) as 
the average Euclidean distance 
between the predicted trajectory and the ground truth 
(i.e., over all the predicted positions of a trajectory) \cite{Pellegrini09, Alahi16}.
The performance evaluation of the proposed intention regularization method is shown in
Table~\ref{tbl:compare}, 
where only the most informative results are shown for a varying probability of intention dropout mask.
As shown in Table~\ref{tbl:compare},
the proposed architecture Lv2 performs better than
Lv1, with the best performance achieved by setting $\gamma=0.3$.

\begin{table}
\centering%
\caption{Comparison of the impact of intention regularization\\ on the APE and the ADE metrics}
\label{tbl:compare}
\begin{tabular}{lc|ccc|c}
\toprule
 &  & \multicolumn{3}{c|}{\footnotesize APE} & {\footnotesize ADE} \\
{\footnotesize Model} & {\footnotesize Mask ($\gamma$)} & {\footnotesize 1h} & {\footnotesize 2h} & {\footnotesize 3h} &   \\
\hline


Lv1\cite{Capobianco2021} & -- & \tablenum{0,57} & \tablenum{	1,14} & \tablenum{ 	1,90} & \tablenum{0,97}  \\

Lv2 (ours)& \tablenum[detect-weight=true]{0}    & \tablenum{0,56} & \tablenum{1,11} & \tablenum{1,86} & \tablenum{0,95} \\

{\bfseries Lv2 (ours)} & {\bfseries \tablenum[detect-weight=true]{0.3}}    & {\bfseries \tablenum[detect-weight=true]{0,51}} & {\bfseries \tablenum[detect-weight=true]{1,03}} & {\bfseries \tablenum[detect-weight=true]{1,78}} & {\bfseries \tablenum[detect-weight=true]{0,88}}  \\
Lv2 (ours)& \tablenum[detect-weight=true]{0.5}    & \tablenum{0,54} & \tablenum{	1,06} & \tablenum{	1,81} & \tablenum{0,91}\\
\bottomrule
\end{tabular}%
\end{table}

\bibliographystyle{IEEEtran}
\bibliography{IEEEabrv,refs}

\end{document}